%% file: Arxiv_V1_Main.tex
\documentclass[12pt]{article}

\usepackage[margin=1in]{geometry}
\usepackage{setspace}
\usepackage{lineno}
\usepackage{microtype}

\usepackage{amsmath,amssymb,mathtools,bm}

\usepackage{graphicx}
\usepackage{booktabs}
\usepackage{siunitx}

\usepackage[square,numbers,sort&compress]{natbib}

\usepackage[hidelinks]{hyperref}

\usepackage{tikz}
\usepackage{subcaption}
\usetikzlibrary{arrows.meta,positioning,fit,calc,backgrounds,decorations.pathreplacing}

\usepackage{enumitem}

\graphicspath{{Figures/}}

\title{\textbf{As Language Models Scale, Low-order Linear Depth Dynamics Emerge}}
\author{
Buddhika Nettasinghe$^{1}$, Geethu Joseph$^{2}$\\
\\
{\small $^{1}$ University of Iowa, Iowa City, USA}\\
{\small $^{2}$Delft University of Technology, Delft, Netherlands}\\
{\small Correspondence to: buddhika-nettasinghe@uiowa.edu}
}
\date{} 

\begin{document}

% Review-style formatting 
\doublespacing
% \linenumbers

\maketitle

% Abstract
\begin{abstract}
\input{sections/00_abstract}
\end{abstract}

\input{sections/01_introduction}
\input{sections/02_related_work}
\input{sections/03_framework_and_identification}
\input{sections/04_results}
\input{sections/05_discussion}

% Methods 
\section*{Methods}
\input{sections/06_methods}

% References 
\bibliographystyle{unsrtnat}
\bibliography{Arxiv_V1_Refs}

\end{document}

%% file: sections/00_abstract.tex
Large language models are often viewed as high-dimensional nonlinear systems and treated as black boxes. Here, we show that transformer depth dynamics admit accurate low-order linear surrogates within context. 
Across tasks including toxicity, irony, hate speech and sentiment, a 32-dimensional linear surrogate reproduces the layerwise sensitivity profile of GPT-2-large with near-perfect agreement, capturing how the final output shifts under additive injections at each layer.
We then uncover a surprising scaling principle: for a fixed-order linear surrogate, agreement with the full model improves monotonically with model size across the GPT-2 family. This linear surrogate also enables principled multi-layer interventions that require less energy than standard heuristic schedules when applied to the full model. Together, our results reveal that as language models scale, low-order linear depth dynamics emerge within contexts, offering a systems-theoretic foundation for analyzing and controlling them.

%% file: sections/01_introduction.tex
\section{Introduction}
\label{sec:introduction}

Large language models are powerful but analytically opaque. Although transformers are organized as a sequence of nonlinear updates to the hidden state across their depth, a systems-level understanding of these updates is still lacking. This gap matters for both mechanism and intervention. Without a tractable model of depthwise response, probing, monitoring and activation-based intervention remain dominated by empirical layer sweeps, fixed schedules and other heuristics. A central question is therefore whether transformer depth dynamics admit a tractable local dynamical description, and whether the quality of that description changes systematically with model scale.

Here we study transformer-based language models from this systems perspective. Treating depth as discrete time and last-token hidden state as the state of the system, we identify a lower-dimensional linear surrogate that approximates how the state propagates through subsequent transformer blocks and influences a final readout in a given context. Consequently, it also reveals how an intervention to the hidden state introduced at a given layer propagates through subsequent blocks and influences the final readout. 

This perspective reveals a striking empirical regularity. Across a diverse task suite, low-order linear surrogates reproduce the layerwise sensitivity profiles of the full nonlinear transformer with high fidelity, capturing not only the best intervention layer but the full shape of the depth response. More surprisingly, identifiability of such a linear surrogate itself improves with model size. At fixed reduced order, agreement between predicted and empirical gain profiles increases monotonically from GPT-2 to GPT-2-medium to GPT-2-large, indicating that larger models admit better linear surrogates. These results collectively support a systems-level empirical regularity: \emph{as transformer language models scale, their local depth dynamics are increasingly well captured by compact low-dimensional linear surrogates.}

The same surrogate also has immediate operational value. Because it is explicit, linear and low-dimensional, it supports principled design objectives such as predicting layerwise gains and choosing multi-layer interventions that minimize energy for a target effect. When transferred back to the full nonlinear transformer, these model-based intervention schedules outperform standard heuristic baselines such as last-layer injection and uniform all-layer injection. 

Together, these results suggest a shift in how large language models can be studied. Even though larger transformers are globally more complex, as they scale, their local depth dynamics are increasingly well approximated by low-dimensional linear models. This raises the possibility that model scale improves not only capability, but also the fidelity of compact mechanistic abstractions. 
Figure~\ref{fig:framework} summarizes this framework. 
A prompt induces a depth-indexed hidden-state trajectory; around that trajectory, the frozen-context dynamics are locally identified and projected to a reduced surrogate; the surrogate then predicts depthwise sensitivity and yields intervention schedules that can be validated directly in the full model.

%% file: sections/02_related_work.tex
\subsection*{Related work}
\label{subsec:related_work}

Recent work on activation steering has shown that linear directions in internal representation space can modify model behavior without retraining. Activation Addition and related activation-engineering methods demonstrated that contrastive activation vectors injected during the forward pass can control properties such as sentiment or topic \cite{Turner2024ActivationEngineering}, and subsequent work showed that such directions can be learned more systematically and transferred across settings \cite{Rimsky2024CAA,Beaglehole2026Universal}. These studies establish that useful internal directions can be identified and applied, but they do not directly address how perturbations introduced along those directions propagate across depth within a given prompt context.

A complementary line of work asks why linear concept directions should exist at all. Park, Choe and Veitch formalized the linear representation hypothesis, clarifying when high-level concepts can be represented linearly and relating this view to probing and intervention \cite{Park2024LRH}. This provides a representational foundation for concept steering, but it is largely static: it explains why linear directions may encode semantics, rather than how perturbations injected along those directions evolve through the network.

Other recent studies suggest that trained transformers exhibit substantial local dynamical regularity. Aubry \emph{et al.} reported aligned Jacobian singular directions, structured residual-stream growth and unexpectedly linear hidden trajectories across large language models \cite{Aubry2024TransformerAlignment}. Such findings support the premise that local linearization of transformer dynamics can be meaningful after training. Here we build on that premise by identifying a low-order surrogate and testing it directly against full-model intervention behavior.

Control-theoretic perspectives have also been taken in language-model literature. Nguyen \emph{et al.} cast activation steering in a feedback-control framework \cite{nguyen2025activation}; Soatto \emph{et al.} study controllability at the prompt level \cite{Soatto2023TamingAIBots}; Moon develops a local-linearization view of neural-network interpretability through controllability, observability and Hankel singular values \cite{Moon2025ControlTheoreticInterpretability}; and Golden suggests that, for a fixed input sequence, LLM next-token inference admits an equivalent linear representation~\cite{golden2025equivalent}. These works motivate a systems perspective, but they do not identify a reduced-order linear dynamical surrogate of depthwise hidden-state response in causal transformers, which is our focus here.

Taken together, prior work provides steering directions, geometric motivation and evidence of local regularity. Our work connects these strands to reveal a broader systems-level empirical regularity: as transformer language models scale, their local depth dynamics are increasingly well captured by much smaller linear surrogates. This frames activation-based intervention as a consequence of an emerging predictive structure in transformer depth dynamics, rather than a heuristic search over layers.

%% file: sections/03_framework_and_identification.tex
\section{A local systems framework for transformer depth dynamics}
\label{sec:framework}

The central idea of this paper is to ask whether the local response of the last-token hidden state can be described by a tractable state-space model that is accurate enough to predict depthwise sensitivity and useful enough to guide intervention design. 

\begin{figure*}[t]
\centering
\includegraphics[width=\textwidth]{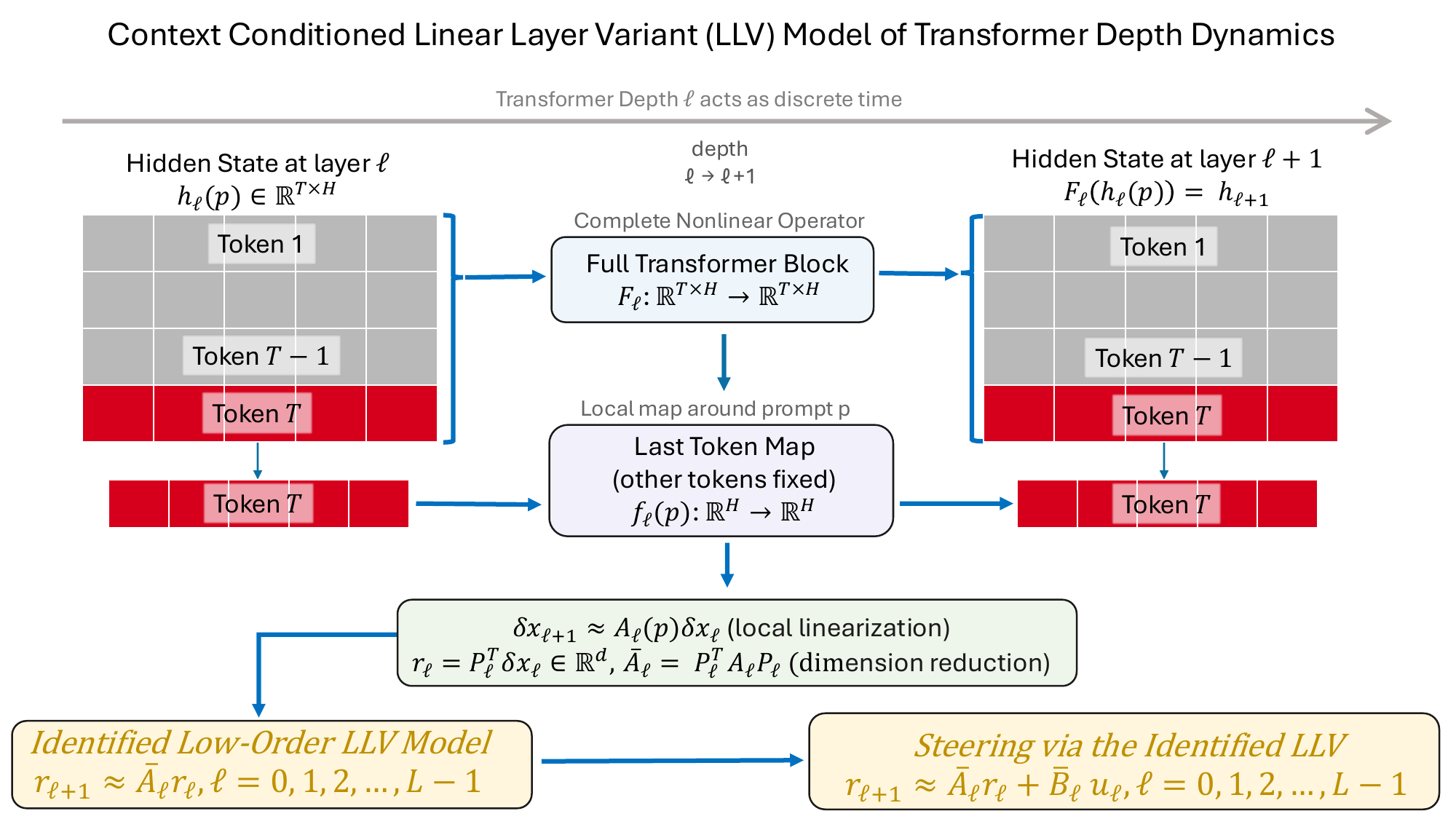}
\caption{Overview of the framework. For a prompt $p$, transformer depth is treated as discrete time and the state is the last non-padding token representation $x_\ell(p)$. Around a prompt-conditioned operating trajectory, the frozen-context last-token map is locally linearized and projected onto a concept-anchored reduced basis to obtain a low-dimensional LLV surrogate. The surrogate predicts layerwise steering gains and yields low-energy multi-layer intervention schedules that are validated in the full transformer.}
\label{fig:framework}
\end{figure*}

Let $h_\ell(p)\in\mathbb{R}^{T\times H}$ denote the hidden-state matrix at depth $\ell$ for prompt $p$, and let $t(p)$ be the index of the last non-padding token. We define the system state as the hidden representation of that token,
\[
 x_\ell(p)=h_\ell(p)[t(p),:]\in\mathbb{R}^H,
\]
so that the state follows the token whose downstream readout is directly affected by activation steering. For a fixed operating prompt, we then freeze all non-last-token rows and vary only the last-token state before the next block. This induces a prompt-conditioned \emph{frozen-context} map
\[
 x_{\ell+1}=f_\ell(x_\ell;p),
\]
which defines the local depth-domain dynamics relevant to steering.

Linearizing this map around the operating trajectory $\bar x_\ell(p)$ yields a prompt-conditioned Jacobian $A_\ell(p)$. Steering is modeled as additive actuation to the last-token residual stream before depth $\ell$ along a depth-indexed concept direction $v_\ell$, estimated from class-conditional mean differences on a separate concept split (see Methods). In the main experiments, actuation is concept-only, so the effective one-step local response is
\[
 \delta x_{\ell+1}\approx A_\ell(p)\,\delta x_\ell + A_\ell(p)v_\ell u_\ell.
\]
To obtain a tractable surrogate, we project these local dynamics onto a reduced basis $P_\ell\in\mathbb{R}^{H\times d}$ whose first column is exactly the concept direction and whose remaining columns are chosen by a reachability-informed Krylov construction. This yields the reduced prompt-conditioned Linear Layer Variant (LLV) model
\[
 r_{\ell+1}\approx \bar A_\ell(p)r_\ell + \bar B_\ell(p)u_\ell,
\qquad
 r_\ell=P_\ell^\top \delta x_\ell,
\]
with
\[
 \bar A_\ell(p)=P_{\ell+1}^\top A_\ell(p)P_\ell,
 \qquad
 \bar B_\ell(p)=P_{\ell+1}^\top A_\ell(p)v_\ell.
\]
The identified matrices are computed by 
% Jacobian--vector products, using forward-mode JVP when available and 
central finite differences.
% otherwise. 
The main results use reduced order $d=32$.

Once identified, the reduced model makes two quantities immediately available. The first is the \emph{predicted layerwise gain curve}. Let $C=v_L^\top P_L$ be the final concept readout in reduced coordinates, and let $\Phi(i,j)$ denote the reduced transition product from depth $i$ to $j$. For a single intervention at layer $k$, the predicted final sensitivity is
\[
 g_k^{\mathrm{pred}}\approx C\,\Phi(k+1,L)\,\bar B_k.
\]
The corresponding empirical gain is measured directly in the full transformer by adding $\pm\epsilon v_k$ before block $k$ and estimating the central finite difference of the final concept score on held-out prompts. Agreement between the predicted and empirical gain curves is our main identifiability criterion.

The second quantity is the \emph{minimum-energy multi-layer control direction}. If controls are stacked across layers, the reduced output shift is linear in the control vector, $\delta y\approx h^\top u$, where the entries of $h$ are the predicted single-layer gains. The minimum-norm control achieving a target concept shift $\Delta y_{\mathrm{tar}}$ is therefore
\[
 u^\star(\Delta y_{\mathrm{tar}})=\frac{\Delta y_{\mathrm{tar}}}{\|h\|_2^2}h.
\]
This closed-form direction is computed from the surrogate and then evaluated in the full nonlinear model by scaling it with a single amplitude and finding the smallest amplitude that reaches the target shift.

This framework turns transformer depth response into a local system-identification problem. Starting from labelled prompts, we estimate concept directions, define prompt-conditioned last-token dynamics, identify a reduced LLV surrogate and use that surrogate to predict where the model is sensitive and how actuation should be distributed across layers. The key question is then empirical: how accurate is this surrogate, how does its identifiability scale with model size, and does it improve intervention design in the full transformer? Those three questions are answered in the next section.

%% file: sections/04_results.tex
\section{Results}
\label{sec:results}

\subsection{LLV surrogates predict steering sensitivity across depth and across tasks}
\label{sec:results_gain_prediction}

\begin{figure*}[t]
\centering
\includegraphics[width=\textwidth]{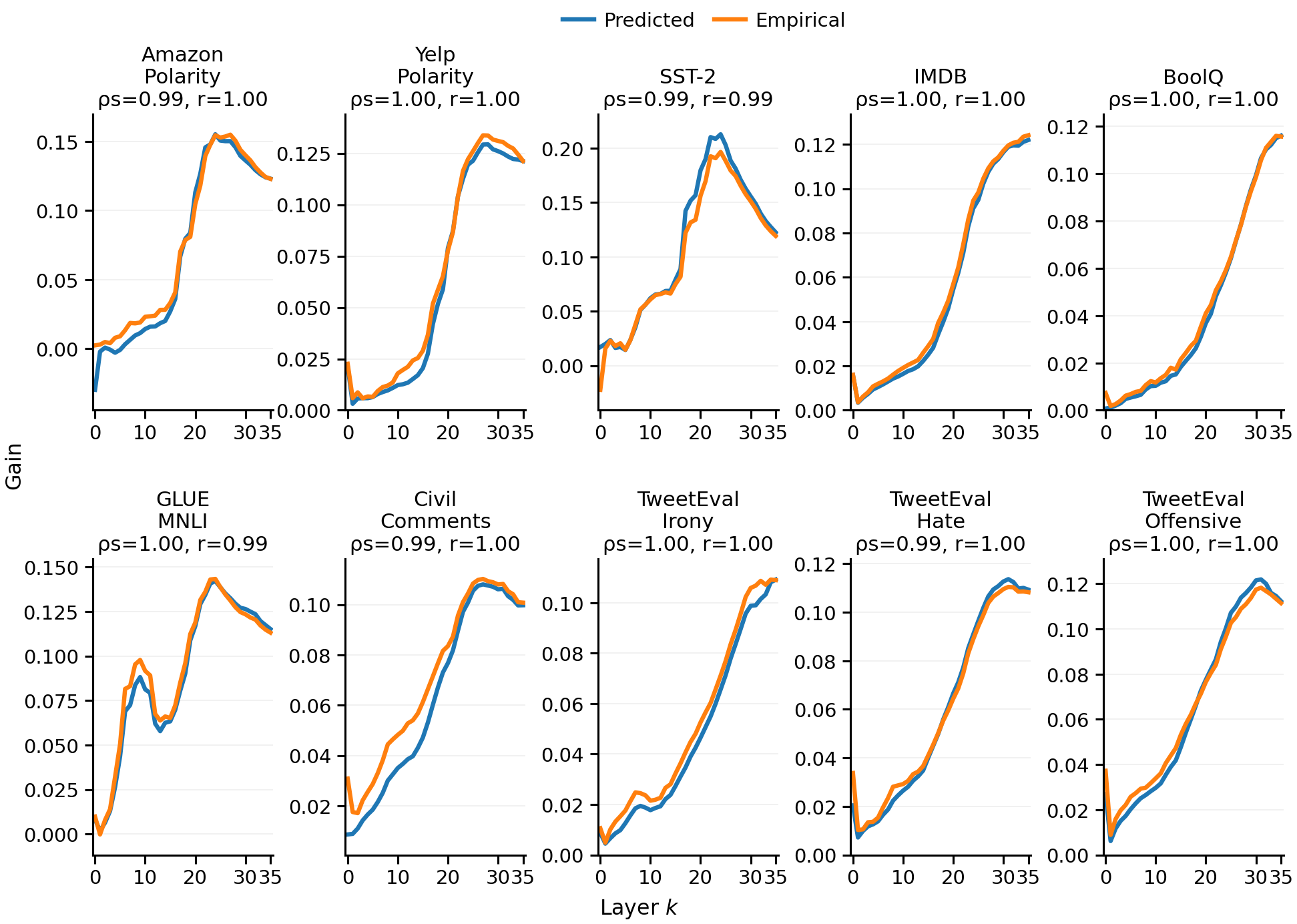}
\caption{Predicted and empirical layerwise steering gains on GPT-2-large at reduced order $d=32$ and evaluation magnitude $\epsilon=0.1$. Each panel shows the gain profile over depth for one dataset. The LLV surrogate predicts the full gain curve rather than only the identity of a best layer.}
\label{fig:gain_curves}
\end{figure*}

A first test of the framework is whether the identified surrogate can predict \emph{where} an internal intervention will be most effective. For each dataset, we therefore compare the layerwise gain profile predicted by the context-conditioned LLV surrogate to the empirical gain profile measured directly in the full transformer by activation addition at each depth. The resulting agreement is unusually strong. In GPT-2-large with reduced order $d=32$, the predicted and empirical gain curves in Fig.~\ref{fig:gain_curves} track each other closely across the entire task suite, with Spearman correlations of $0.99$ or $1.00$ in all displayed datasets.

This is a demanding test of identifiability. The gain at a given layer depends on how an injected perturbation is transformed by all downstream blocks, so accurate prediction requires a faithful model of local depth propagation rather than merely a plausible steering direction. The agreement in Fig.~\ref{fig:gain_curves} shows that, within context, the LLV surrogate captures the dominant mechanism governing steering sensitivity across depth. It predicts not only high-gain layers, but the overall \emph{shape} of the response, including smooth late-depth ramps, broad middle-to-late plateaus and non-monotone gain landscapes.

Several qualitative features of Fig.~\ref{fig:gain_curves} are important. First, the optimal intervention depth is clearly task-dependent. Some tasks exhibit almost monotone late-layer amplification, whereas others show broad peaks or extended plateaus across a band of middle and late layers. This immediately argues against a universal heuristic such as always steering at the final block. Second, in the strongest cases the predicted and empirical profiles are nearly visually indistinguishable over the full network depth. Third, cases in which exact top-$k$ layer overlap would appear ambiguous are often those with wide high-gain plateaus, where many neighbouring layers are effectively tied. For that reason, we treat Pearson and Spearman agreement as the primary metrics and use top-$k$ overlap only as a secondary diagnostic.

The gain-prediction result is also robust to the main approximation choices. In Extended Data Fig.~2, agreement remains high over a broad range of perturbation magnitudes $\epsilon$, indicating that the empirical finite-difference estimator is operating in a stable local-response regime rather than at a finely tuned scale. Agreement also improves as the reduced dimension is increased from very small values and then saturates, showing that the steering-relevant local dynamics are compact. Finally, replacing a random complement with a reachability-informed Krylov complement systematically improves prediction, especially in the harder cases, consistent with the idea that steering effects are concentrated in a low-dimensional reachable subspace.

These results establish the first main finding of the paper. Context-conditioned LLV surrogates do not merely provide a post hoc description of activation steering; they predict the depthwise sensitivity structure of the full nonlinear transformer with high fidelity. This turns the question of where to intervene from a brute-force layer sweep into a predictive systems problem.

\subsection{A scaling principle: larger transformers admit better low-order LLV surrogates}
\label{sec:results_scaling}

\begin{figure*}[t]
\centering
\includegraphics[width=0.49\textwidth]{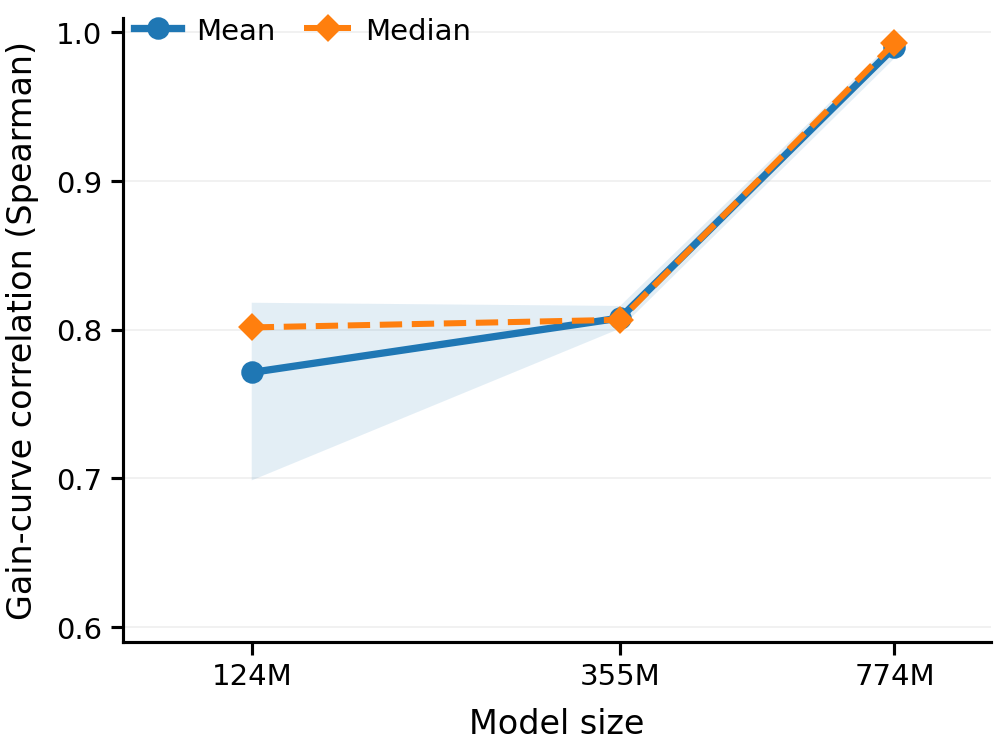}\hfill
\includegraphics[width=0.49\textwidth]{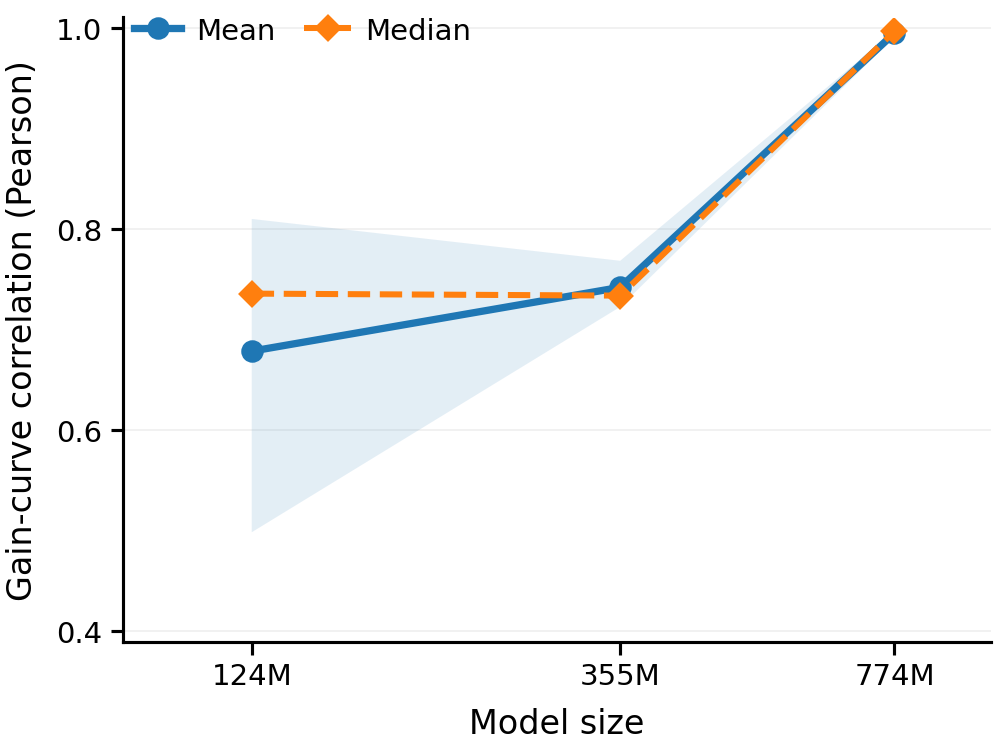}
\caption{Scaling of LLV identifiability at fixed reduced order $d=32$ and evaluation magnitude $\epsilon=0.1$. Aggregate Spearman and Pearson agreement increase monotonically from GPT-2 to GPT-2-medium to GPT-2-large. By GPT-2-large, the identified low-order surrogate nearly saturates the gain-prediction task.}
\label{fig:scaling_main}
\end{figure*}

The most surprising scientific finding is that identifiability itself improves with model scale. Figure~\ref{fig:scaling_main} summarizes the scaling experiment at fixed reduced order $d=32$ across the GPT-2 family. Aggregate prediction quality increases monotonically from GPT-2 to GPT-2-medium to GPT-2-large in both Spearman and Pearson agreement. Mean Spearman rises from about $0.77$ in GPT-2 to $0.81$ in GPT-2-medium and $0.995$ in GPT-2-large, while mean Pearson rises from about $0.68$ to $0.74$ to $0.997$. By GPT-2-large, the low-order surrogate is nearly saturated.

This is not simply the statement that larger models are easier to steer. It is the stronger systems statement that larger models are more cleanly \emph{identifiable} through a compact local dynamical surrogate. Under the same reduced-order budget, the steering-relevant depth response of the larger model is better captured by a small state-space abstraction. Stated differently, scale appears to make the local response structure not only more capable, but more compressible and more predictable.

The by-dataset scaling panels in Extended Data Fig.~3 show that the monotone trend is strongest at the aggregate level rather than uniformly task-by-task. Some datasets are already highly predictable in smaller models, while others improve sharply only at larger scale. Nevertheless, the overall trend is unmistakable: both rank-based and linear agreement improve substantially with model size, and GPT-2-large exhibits near-perfect gain-profile prediction across the entire suite. The pattern is especially striking because a larger model is globally more complex, yet locally more faithfully described by a compact surrogate.

A plausible systems intuition is that increasing width and representational redundancy stabilise local Jacobian responses around operating trajectories and reduce estimation variance in the projected dynamics. Better-formed concept directions may also align more strongly with a small number of dominant reachable modes, making low-order reduction more faithful. Our results do not prove this mechanism, but they reveal an empirical regularity that is surprising on its own: within the tested scaling family, larger transformers admit better low-order LLV surrogates.

This scaling principle is important because it elevates the contribution beyond a single-model steering method. It suggests that model scale changes the quality of the local systems abstraction itself. If this regularity extends across additional model families, identifiability may become a useful systems-level quantity for comparing architectures, training regimes and intervention strategies.

\subsection{Design payoff: minimal-energy multi-layer control outperforms heuristic schedules}
\label{sec:results_control}

\begin{figure*}[t]
\centering
\includegraphics[width=\textwidth]{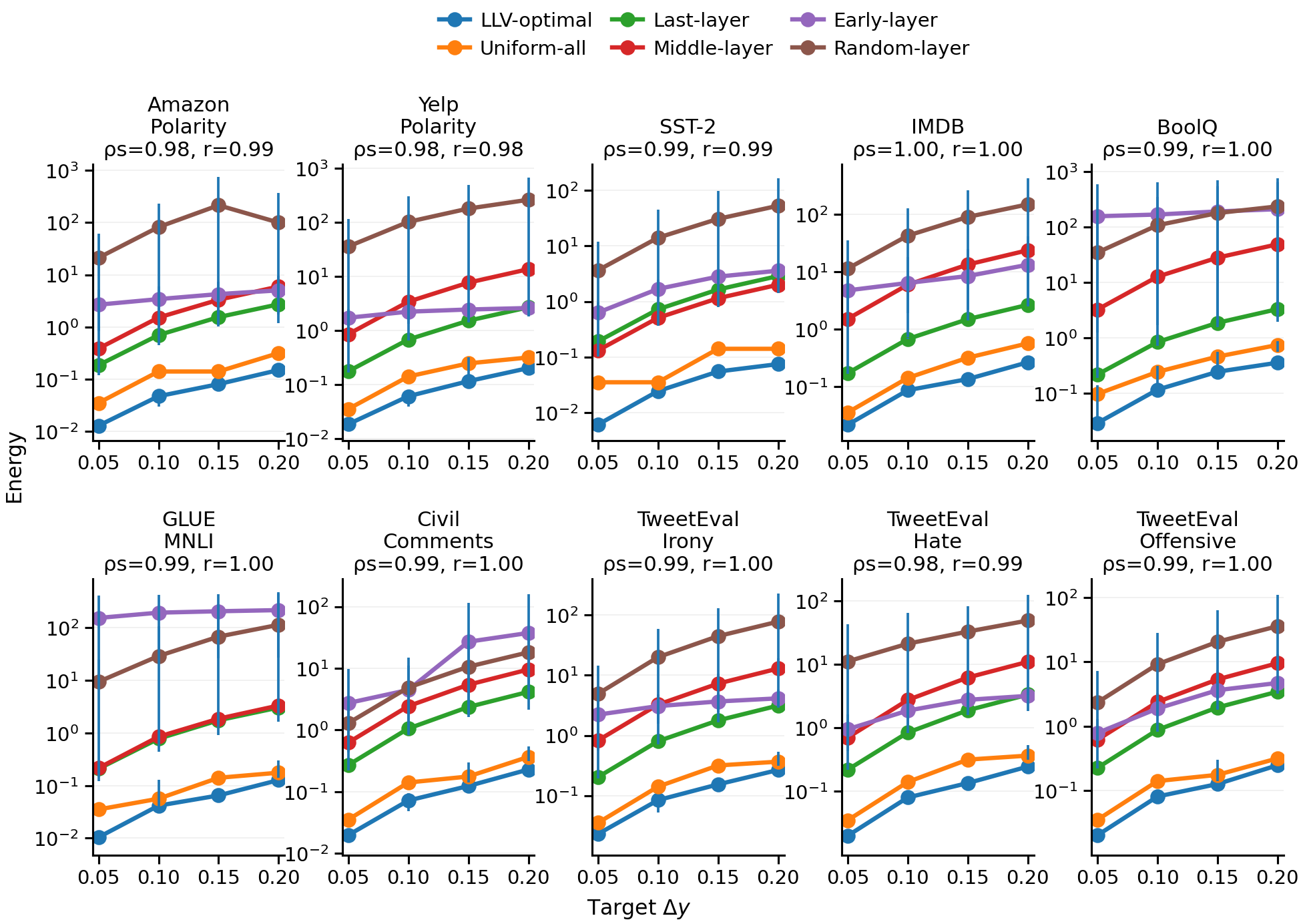}
\caption{Minimal-energy steering on GPT-2-large. Left, energy required to reach target concept shifts $\Delta y$ for the LLV-optimal schedule and heuristic baselines. Right, the same quantities normalised by the LLV-optimal energy. The LLV-derived policy is consistently lowest-energy or tied-lowest-energy across datasets and target shifts, while uniform injection is the strongest non-model baseline and single-layer or random schedules are often much worse.}
\label{fig:control_main}
\end{figure*}

A useful surrogate should do more than reproduce sensitivity curves; it should improve intervention design. We therefore ask a complementary question: given a desired change in the final concept score, how should actuation be distributed across layers to minimise total injection energy? The identified LLV model yields a direct answer. Because the reduced output shift is linear in the stacked control sequence, the minimum-energy multi-layer direction has a closed-form solution, which we term \emph{LLV-optimal}. We then evaluate that direction in the full transformer by scaling it with a single amplitude and finding, by bisection, the smallest amplitude required to reach a target $\Delta y$.

Figure~\ref{fig:control_main} shows the result on GPT-2-large. Across all datasets and all target shifts, the LLV-optimal policy is consistently the lowest-energy or tied-lowest-energy method. The strongest heuristic baseline is typically \emph{uniform-all}, which injects equally across all layers. Even this baseline generally requires more energy than the LLV-derived schedule, often by a factor of roughly two to five. Single-layer schedules are substantially less efficient, with last-layer-only interventions frequently an order of magnitude worse and early-layer or random single-layer policies often worse by one to several orders of magnitude.

This result matters because it closes the loop from explanation to action. The same identified depth response that predicts the gain curves also determines how control should be distributed across layers to reach a target with minimal energy. The surrogate is therefore not merely diagnosing sensitivity after the fact; it is computing a policy that transfers back to the full nonlinear transformer and outperforms commonly used heuristic schedules when tested directly on the original model.

The qualitative pattern across datasets is also informative. When gain is sharply concentrated in a narrow region of depth, the LLV-optimal policy focuses actuation there while still exploiting nearby support where beneficial. When gain is distributed across a broad plateau, the optimum spreads control over multiple layers, whereas single-layer strategies waste energy by over-committing to one location. Uniform-all performs better than arbitrary one-layer baselines because it partially hedges against uncertainty, but it remains suboptimal because it allocates effort to layers that the surrogate correctly predicts to be inefficient. Thus the control results provide a particularly clean operational interpretation of Fig.~\ref{fig:gain_curves}: predicting where the model is sensitive is valuable because it translates directly into lower-energy intervention.

Together, Figs.~\ref{fig:gain_curves}--\ref{fig:control_main} show that the LLV surrogate is not only an accurate local description of transformer depth dynamics within context, but also a useful design object. It predicts where to intervene, becomes more identifiable as models scale, and yields materially better actuator schedules than heuristic alternatives.

%% file: sections/05_discussion.tex
\section{Discussion}
\label{sec:discussion}

The central result of this work is that transformer depth response admits low-order local linear surrogates whose fidelity improves with model scale. Within a fixed context, a causal transformer can be approximated by a low-order Linear Layer Variant (LLV) model that captures how perturbations to internal state propagate across depth and influence a final concept-relevant readout. This is already notable at the level of predictive accuracy: the surrogate reproduces full layerwise sensitivity profiles rather than only identifying a single effective intervention layer. More broadly, it suggests that transformer depth dynamics are locally more structured and more compressible than a purely black-box view would suggest.

The most surprising finding is the scaling regularity. At fixed reduced order, identifiability improves monotonically with model size across the GPT-2 family, indicating that larger models are more accurately captured by low-order local linear surrogates in context. This suggests that model scale changes the quality of the local systems abstraction itself. Although larger transformers are globally more complex, their local depth response in context can become more predictable under the same reduced-order budget. This raises the possibility that scale improves not only capability, but also the fidelity of compact mechanistic descriptions.

This perspective has a broader implication for how large language models are studied. Much current work treats scale as increasing both capability and opacity simultaneously. Our results imply that greater scale may also induce more regular internal dynamics. If that pattern extends across additional model families, then identifiability may become a useful systems-level quantity for comparing architectures, training procedures and intervention paradigms. In that sense, the present results point toward a more general scientific question: whether increasing scale systematically improves the quality of low-dimensional mechanistic abstractions for complex neural systems.

An immediate benefit of identifying a reduced-order LLV model is the ability to determine how actuation inputs should be distributed across layers to achieve a target change in the final readout with minimal energy. The resulting intervention policies outperform standard heuristic schedules (such as last-layer or all-layer actuation) when applied to the full model. Thus, principled intervention design becomes a direct operational consequence of a more fundamental predictive structure in transformer depth dynamics.

The surrogate introduced here should nevertheless be understood as deliberately local. It is identified around prompt-conditioned operating trajectories and is not intended as a global law of transformer computation. Its purpose is to provide a tractable state-space model within a local regime, where it can predict depthwise sensitivity and support principled actuator design, after which the resulting policy is validated in the full nonlinear model. This design-then-validate workflow is therefore essential. The contribution is not to replace a transformer by a linear system in general, but to show that a local linear systems lens can be accurate enough to make analysis and intervention substantially less heuristic.

%% file: sections/06_methods.tex
\subsection*{Models and datasets}

We evaluate causal GPT-2 family models at multiple scales and report detailed gain-prediction and control results on GPT-2-large. The main scaling analysis at reduced order $d=32$ uses GPT-2, GPT-2-medium and GPT-2-large, allowing identifiability to be compared under a common identification pipeline and a fixed reduced-order budget. The primary gain-curve and control figures focus on GPT-2-large, where the LLV approximation is strongest and the control comparison is most informative. The run scripts used for the main figures configure GPT-2-large control runs with concept batch size $32$, held-out batch size $64$, finite-difference step $2\times10^{-3}$, Krylov basis construction from $128$ operating prompts, Krylov complement size $31$, reduced dimension $32$, and gain-evaluation magnitudes $\epsilon\in\{0.05,0.1\}$, with $\epsilon=0.1$ used for the main paper figures.

We use ten binary Natural Language Processing (NLP) tasks spanning sentiment, question answering, textual entailment, toxicity and social-media classification: Amazon Polarity \cite{zhang2015character}, Yelp Polarity \cite{zhang2015character}, SST-2 \cite{socher2013recursive}, IMDB \cite{maas2011learning}, BoolQ \cite{clark2019boolq}, a binary version of GLUE MNLI \cite{williams2018broad,wang2018glue}, Civil Comments Toxicity \cite{borkan2019nuanced}, TweetEval-Irony \cite{barbieri2020tweeteval,vanhee2018semeval}, TweetEval-Hate \cite{barbieri2020tweeteval,basile2019semeval}, and TweetEval-Offensive \cite{barbieri2020tweeteval,zampieri2019semeval}. The objective in each case is to define a concept score whose change under intervention can be measured consistently across prompts.

\subsection*{Prompt construction and data splits}

Prompt templates follow the dataset builders in the experimental setup. Review-style datasets (IMDB, Yelp, Amazon and SST-2) use a shared sentiment wrapper of the form \texttt{Review: \{text\} Overall, I feel}, with Amazon reviews formatted as title-plus-content before wrapping. BoolQ uses a task-native prompt of the form \texttt{Passage: ... Question: ... Answer:}. GLUE MNLI is formatted as premise--hypothesis pairs ending in \texttt{Relationship:}. Civil Comments and TweetEval tasks use the underlying text field directly.

For every dataset, we form three disjoint balanced prompt sets: a \emph{concept} split used to estimate per-layer concept directions, an \emph{operating} split used to identify prompt-conditioned local LLV dynamics, and a \emph{held-out evaluation} split used only for empirical gain and control measurements. The experimental setup enforces class balance at each stage, with nominal sizes $n_{\mathrm{concept}}=400$, $n_{\mathrm{operating}}=200$ and $n_{\mathrm{eval}}=200$ per class where data availability permits. The use of separate operating and evaluation splits is important: predicted gains are not scored on the same prompts used for local identification.

\subsection*{Frozen-context local dynamics and concept directions}

Let $h_\ell(p)\in\mathbb{R}^{T\times H}$ denote the hidden-state matrix at depth $\ell$ for prompt $p$, and let $x_\ell(p)=h_\ell(p)[t(p),:]\in\mathbb{R}^H$ be the hidden state of the last non-padding token. Depth is treated as discrete time. For each prompt in the operating set, we first run the unperturbed model to obtain the depth-indexed operating trajectory. We then define a frozen-context last-token map by holding all non-last-token rows fixed and varying only the last-token state before the next transformer block. This yields a prompt-conditioned local map $x_{\ell+1}=f_\ell(x_\ell;p)$.

Layerwise concept directions are estimated from a separate concept split. For each depth $\ell$, we compute class-conditional means of the last-token state and define the normalized mean-difference direction $v_\ell$. Steering in the present paper is concept-only, so the actuation basis at depth $\ell$ is $V_\ell=v_\ell$.

\subsection*{Reduced LLV identification and Krylov basis construction}

We linearise the frozen-context map around the operating point $\bar x_\ell(p)$, producing a local Jacobian $A_\ell(p)$. To reduce dimensionality, we construct a concept-anchored basis $P_\ell\in\mathbb{R}^{H\times d}$ whose first column is exactly $v_\ell$ and whose remaining columns form a complement. The main results use a reachability-informed Krylov complement rather than a random orthogonal complement. Operationally, we build this complement from mean Jacobian actions across a subset of operating prompts. At depth $\ell$, the post-injection seed direction $A_\ell(p)v_\ell$ is averaged over prompts, propagated forward through mean Jacobian operators, projected orthogonally to the next concept direction, and orthonormalized. The resulting sliding product-Krylov basis privileges directions actually excited by steering perturbations and gives a more faithful reduced model than a random complement.

The reduced prompt-conditioned LLV surrogate is
\[
 r_{\ell+1}\approx \bar A_\ell(p)r_\ell + \bar B_\ell(p)u_\ell,
 \qquad
 r_\ell=P_\ell^\top\delta x_\ell,
\]
with
\[
 \bar A_\ell(p)=P_{\ell+1}^\top A_\ell(p)P_\ell,
 \qquad
 \bar B_\ell(p)=P_{\ell+1}^\top A_\ell(p)V_\ell.
\]
These matrices are computed using Jacobian--vector products (JVPs) without explicitly forming full Jacobians. The implementation first attempts forward-mode JVP and falls back to central finite differences when necessary or when finite-difference-only mode is requested. Main figures are reported for reduced order $d=32$ with Krylov complement size $31$.

\subsection*{Gain curves and agreement metrics}

For a single intervention at layer $k$, the reduced model predicts a final output sensitivity
\[
 g_k^{\mathrm{pred}}\approx C\,\Phi(k+1,L)\,\bar B_k,
\]
where $\Phi$ is the reduced depth transition and $C$ is the final concept readout in reduced coordinates. The implementation computes these predicted gains for each operating prompt and averages the resulting gain curves across prompts. This order of operations matters because products of transition matrices are nonlinear in the reduced Jacobians.

Empirical gain curves are measured in the full transformer by adding $\pm\epsilon v_k$ to the last-token activation before block $k$, evaluating the final concept score on held-out prompts, and estimating the derivative-like gain by central difference,
\[
 g_k^{\mathrm{emp}}(\epsilon)=\frac{\mathbb{E}[y^{(+\epsilon,k)}]-\mathbb{E}[y^{(-\epsilon,k)}]}{2\epsilon}.
\]
The main gain-curve figures are shown at $\epsilon=0.1$, while robustness sweeps vary $\epsilon$ over $\{0.01,0.02,0.05,0.1,0.2,0.3,0.4,0.5\}$. Agreement between predicted and empirical curves is quantified primarily by Spearman rank correlation and Pearson correlation. Top-$k$ overlap is treated only as a secondary diagnostic because broad high-gain plateaus can make exact set overlap unstable even when the full gain profile is predicted accurately.

\subsection*{Scaling aggregation}

Scaling figures are produced by running the same identification pipeline on GPT-2, GPT-2-medium and GPT-2-large, using the same dataset suite, the same reduced order $d=32$, the same Krylov construction and the same evaluation magnitude $\epsilon=0.1$. For each model and random seed, Spearman and Pearson agreement are computed separately for each dataset. Aggregate scaling summaries are then formed by averaging correlations across datasets within each seed and reporting mean and median across seeds. The figure-generation script also writes the by-dataset rows used for the per-dataset scaling panels in Extended Data Fig.~3.

\subsection*{Minimal-energy multi-layer control}

For multi-layer concept-only actuation, we allow simultaneous injections
\[
 x_k \leftarrow x_k + u_k v_k,\qquad k=0,\dots,L-1,
\]
and measure control energy by the squared $\ell_2$ norm of the coefficient vector,
\[
 E(u)=\sum_{k=0}^{L-1}u_k^2=\|u\|_2^2.
\]
In the reduced surrogate, the final concept shift is linear in the stacked control sequence,
\[
 \delta y\approx h^\top u,
\]
where the entries of $h$ are precisely the predicted single-layer gains. The minimum-energy control for a target shift $\Delta y_{\mathrm{tar}}$ is the minimum-norm solution of
\[
 \min_u \|u\|_2^2 \quad \text{subject to} \quad h^\top u=\Delta y_{\mathrm{tar}},
\]
which yields
\[
 u^\star(\Delta y_{\mathrm{tar}})=\frac{\Delta y_{\mathrm{tar}}}{\|h\|_2^2}h.
\]
In practice we compute the associated direction $u_{\mathrm{dir}}\propto h$, orient it so that a small probe actuation increases the concept score, and then evaluate it in the full transformer by scaling with a single amplitude $\alpha$. For each target $\Delta y\in\{0.05,0.1,0.15,0.2\}$, one-dimensional bisection is used to find the smallest amplitude whose realized full-model shift reaches the target. The reported minimal energy is therefore
\[
 E^\star(\Delta y)=\|\alpha^\star u_{\mathrm{dir}}\|_2^2.
\]
The main baselines are uniform injection across all layers, last-layer-only, middle-layer-only, early-layer-only and random single-layer injection. Energy-ratio plots normalize all baselines by the LLV-optimal energy at the same target.

\subsection*{Compute and reproducibility}

All experiments use HuggingFace transformer implementations with local Jacobian actions estimated by JVPs or finite differences. The suite runner organizes outputs by model and seed, saving gain metrics, optional gain arrays, control CSVs and figure-ready summaries. The scaling analysis uses three random seeds for GPT-2, GPT-2-medium and GPT-2-large. The GPT-2-large gain-curve visualization in Fig.~\ref{fig:gain_curves} is shown for seed 0, while Fig.~\ref{fig:scaling_main} aggregates over seeds. Control figures are generated from the completed GPT-2-large control runs available in the current suite.

The identification and evaluation process uses padding-safe last-token indexing, concept-balanced splits and fixed prompt templates so that operating and evaluation conditions are comparable across runs. The experiments produced the main results shown in 
Fig.~\ref{fig:gain_curves}, Fig.~\ref{fig:scaling_main} and Fig.~\ref{fig:control_main}.
Experimental configuration files, figure-generation scripts and processed outputs for the main results will be released to support full reproducibility.